\documentclass[review]{elsarticle}

\usepackage{lineno,hyperref}
\usepackage{subfigure}
\usepackage{textcomp}
\usepackage{amsmath}
\usepackage{placeins}
\usepackage{multirow}
\usepackage{float}
\usepackage{layouts}
\usepackage{graphicx}
\usepackage{printlen}
\usepackage{parskip}

\journal{}

\begin{document}

\begin{frontmatter}

\title{Firefly Algorithm for optimization problems with non-continuous variables: A Review and Analysis}

\author{Surafel Luleseged Tilahun and Jean Medard T Ngnotchouye}
\address{School of Mathematics, Statistics and Computer Science, \\ University of KwaZulu-Natal}


\cortext[mycorrespondingauthor]{Corresponding author}
\ead{surafelaau@yahoo.com}


\begin{abstract}
Firefly algorithm is a swarm based metaheuristic algorithm inspired
by the flashing behavior of fireflies. It is an effective and an
easy to implement algorithm. It has been tested on different
problems from different disciplines and found to be effective. Even
though the algorithm is proposed for optimization problems with
continuous variables, it has been modified and used for problems
with non-continuous variables, including binary and integer valued
problems. In this paper a detailed review of this modifications of
firefly algorithm for problems with non-continuous variables will be
discussed. The strength and weakness of the modifications along with
possible future works will be presented.
\end{abstract}


\end{frontmatter}


\section{Introduction}\label{sec1}

Optimization problems are problems of finding values for the
variables which will give an optimum functional value of the
objective function. This kind of problems exists beyond our daily
activity. They are common problems, in engineering, decision
science, agriculture, computer science, economics and many other
disciplines \cite{Pike14, Tilahun12, Hamadneh12, Ong11, Tilahuns12,
Lucia90, Joshi13, TilahunO12, Parolo09, Shimoyam11}. Based on the
decision variables a solution can be classified into three
categories as continuous variable, non-continuous variables and
mixed variables. Continuous variables are when the variable can have
any value in the given interval and if that is not the case then we
have a problem with non-continuous variables which includes integer
and binary variables. When some of the decision variables can be
assigned with continuous and the rest with non-continuous values
then it is called a mixed problem. Most of real optimization
problems involves non-continuous variables like number of products
and human resource.

For an optimization problem there are different solution methods.
One class of solution methods is metaheuristic algorithms. These
algorithms are a non-deterministic solution method which search the
solution space based on an educated guess and 'trail and error'
approach based on a given randomness term. Swarm based algorithms
are a class of metaheuristic optimization algorithms which are
inspired by the social behavior of animals. Firefly algorithm is one
of swarm based metaheuristic algorithm inspired by the flashing
behavior of flashing bugs also called fireflies \cite{Yang08}.
Firefly algorithm is an easy to implement algorithm which can easily
be implemented and can also easily be parallelized. It is also
tested to be effected on problems from different problem domain.
Even though it has a number of strength it also prone to parameter
setting and also on controlling the exploration and exploitation of
the search space. Hence, different modified versions are proposed to
improve its performance as well as to make it useful for problems
with non-continuous variables.

Hence in this paper, the modifications proposed to make firefly
algorithm suitable for optimization problems with non-continuous
variables will be discussed. A discussion on the strength and
weakness of the modifications will be presented along with possible
future work. In the next section a general discussion on
optimization problems will be given followed by a discussion on
firefly algorithm. In section 3, a discussion on modified versions
of firefly algorithms will be given followed by a general discussion
on the modifications in section 4. In section 5 conclusion will be
presented.

\section{Preliminary}

\subsection{Optimization problems}

A given optimization problem has decision variables $x=(x_1, x_2, .
. ., x_n)$, for which we are search a value for, an objective
function, $f(x)$, which is a function of the decision variable and
also used to measure the performance of the values assigned to the
decision variables and a feasible region, $S$, from which the
decision variable can take values. A minimization problem can then
be given as in equation \eqref{eq1}

\begin{equation}
\min\limits_x\{f(x)|x\epsilon S \subseteq \Re^n \} \label{eq1}
\end{equation}

The search space or the feasible region $S$ can be continuous,
non-continuous or mixed i.e. continuous for some of the variables
and non-continuous for the others possibly binary or integer.

A solution $x^*$ is said to be global (local) optimal solution for
the minimization problem given in equation \eqref{eq1} if and only
if $x \epsilon S$ and $f(x^*) \leq f(x)$ for all $x \epsilon S$ (for
all $x$ in the neighborhood of $x^*$).

\subsection{Firefly algorithm}
Nature has been a motivation to different metaheuristic
algorithms\cite{Yang08, Babaoglu07, Miao14}. The interaction between
a predator and its pray, the attraction between bodies and the swarm
behavior of fishes or birds can be mentioned as an example
\cite{Dorigo04, PSO, TilahunPPA15, Yuce13}. Firefly algorithm is
inspired by the flashing behavior of flashing bugs also called
fireflies. It is proposed for optimization problems with continuous
variables \cite{Yang08}. A randomly generated feasible solutions
will be considered as fireflies where their brightness is determined
by their performance on the objective function. The algorithm is
guided by three rules. The first rule is that fireflies are unisex,
that means any firefly can be attracted to any other firefly. The
second rule is the brightness of a firefly depends on its
performance in the objective function. The attraction of a firefly
depends on its brightness and decreases with distance. This means,
since we are considering a minimization problem, a solution with
smaller functional value is brighter. The light intensity has an
inverse square law as given in equation \eqref{eq2}.

\begin{equation}
I\prec \frac{1}{r^2}\label{eq2}
\end{equation}

where $I$ is the intensity and $r$ is the distance. The brightness
follows similar rule as the light intensity with respect to the
distance. Furthermore, suppose the light is passing through a medium
with a light absorbtion coefficient of $\gamma$. The the brightness
of a firefly at a distance $r$ can be summarized using the equation
given in equation \eqref{eq3}'

\begin{equation}
\beta = \beta_0e^{-\gamma r^2} \label{eq3}
\end{equation}

where $\beta$ is the brightness of the firefly at a distance $r$ and
$\beta_0$ is the brightness at the source, i.e. $r=0$.

A solution $x_i$ will be attracted by a brighter firefly $x_j$, this
means $x_i$ moves towards $x_j$ using \eqref{eq4}.

\begin{equation}
x_i:=x_i+\beta_0e^{-\gamma r_{ij}^2}(x_j-x_i)\label{eq4}
\end{equation}

In addition it will explore using a random movement given by
\eqref{eq5}.

\begin{equation}
x_i:=x_i+\alpha (rand()-0.5) \label{eq5}
\end{equation}

for a step length for the random movement $\alpha$ and $rand()$ is
an $n$ vector whose entries are generated randomly from a uniform
distribution between zero and one.

Therefor, for the updating equation of a firefly $x_i$, by putting
equations \eqref{eq4} and \eqref{eq5}, is given as in equation
\eqref{eq6}.

\begin{equation}
x_i:=x_i+\beta_0e^{-\gamma r_{ij}^2}(x_j-x_i)+\alpha
(rand()-0.5)\label{eq6}
\end{equation}

for a brighter firefly $x_j$. If there is no brighter firefly than
$x_i$ it will perform a random move only as given in equation
\eqref{eq5}.

The algorithm is summarized in Table \ref{Table1}.

\begin{table}[H]
\centering \caption {The standard firefly algorithm} \centering
\begin{tabular}{l}
 \hline
 Set algorithm parameters ($\gamma, \alpha$)\\
 Set simulation set-up (Maximum number of iteration ($MaxGen$), \\
  \; \; \; \; \; \; \; \; \; \; \; \; \; \; \; \; \; \;  \; \; \; \; \; \; \; \; \;Number of initial solutions ($N$))\\
 Randomly generate $N$ feasible solutions ($x_1, x_2, . . ., x_N$)\\
 for $iteration=1:MaxGen$\\
  \; \; \, Compute the brightness\\
  \; \; \, Sort the solutions in such a way that $I_i \geq I_{i-1}$, $\forall i$ \\
  \; \; \, for $i=1:n-1$ \\
  \; \; \, \; \; \, for $j=i+1:n$ \\
  \; \; \, \; \; \, \; \; \, if ($I_i>I_j$)\\
  \; \; \, \; \; \, \; \; \, \; \; \, move firefly $i$ towards firefly $j$\\
  \; \; \, \; \; \, \; \; \, end if \\
  \; \; \, \; \; \, end for \\
  \; \; \, end for \\
  \; \; \, move firefly $N$ randomly\\
  end for \\\hline
\end{tabular}\label{Table1}
\end{table}

\section{Modified firefly algorithms for discrete optimization problems}

Firefly algorithms has been modified and used for discrete
optimization problems. Based on the space where the updating is
performed, these modifications can generally be categorized in to
two categories. The fist category is when the updating is done in
the continuous space and a discretization mechanism is used to
change the values to discrete numbers whereas the second is when the
updating is done on the discrete space.

\subsection{Updating in the continuous space}
In this category the updating procedure of the standard firefly
algorithm will be used and the result will be converted to discrete
values.

\begin{figure}[H]
\centering
\includegraphics[width=10.5cm, height=9cm]{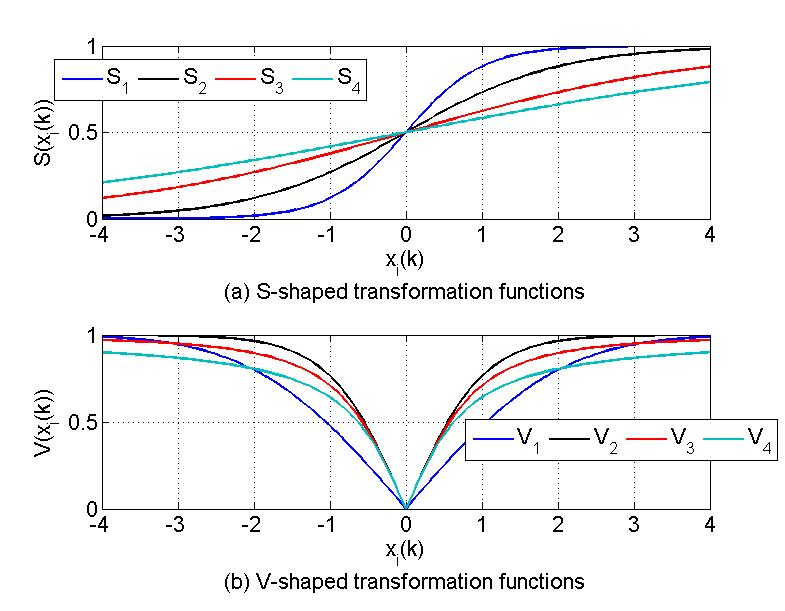}
\caption{S-shape and V-shape transfer functions.}\label{fig1}
\end{figure}

Perhaps the first modification of firefly algorithm for discrete
problems, specifically binary problems, is done for a job scheduling
problem in \cite{Sayadi104}. After updating a solution $x_i$ using
the updating equations of the standard firefly algorithm, each
component $k$ of the solution vector $x_i$ will be converted to 0's
and 1's based on the sigmoid function,
$S(x_i(k))=\frac{1}{1+e^{-x_i(k)}}$, as the probability that
$x_i(k)$ will be 1. A similar approach in which
$x_i(k)=\{\begin{array}{l}
          1, rand < S(x_i(k)) \\
          0, otherwise
          \end{array}$ is used in \cite{Palit106, Yang98, Sayadi105, Rajalakshmi100, Chhikara99}.
In addition using the sigmoid function the tan hyperbolic function
has also be used in some studies \cite{Chandrasekaran107,
Chandrasekaran97, Crawford96}. In \cite{Crawford96}, rather than
using a random number to determine the value of $x_i(k)$, a given
threshold, $\tau$, is used, i.e. $x_i(k)=\{\begin{array}{l}
          1, \tau < tan|x_i(k)| \\
          0, otherwise
          \end{array}$. Furthermore, in \cite{Liu108} both the sigmoid
as well as the tan hyperbolic functions are used to limit the value
of $x_i(k)$ in between 0 and 1. The two functions are shown on Fig.
\ref{fig1}, where the sigmoid function falls as an S-shaped function
and the tan hyperbolic function as a V-shaped function.

In \cite{Crawford95}, four variants of the sigmoid functions,
S-shaped functions, and four variants of the tan hyperbolic
functions, V-shaped functions are given. The four S-shaped functions
are given by $S_1(x_i(k))=\frac{1}{1+e^{-2x_i(k)}}$,
$S_2(x_i(k))=\frac{1}{1+e^{-x_i(k)}}$,
$S_3(x_i(k))=\frac{1}{1+e^{-\frac{x_i(k)}{2}}}$ and
$S_4(x_i(k))=\frac{1}{1+e^{-\frac{x_i(k)}{3}}}$ and the four
V-shaped functions are
$V_1(x_i(k))=|erf(\frac{\sqrt{2}}{\pi}x_i(k))|$,
$V_2(x_i(k))=tanh(|x_i(k)|)$, $V_3(x_i(k))=|\frac{x}{\sqrt{1+x^2}}|$
and $V_4(x_i(k))=|\frac{2}{\pi}arctan(\frac{\pi}{2}x_i(k))|$. The
transformation functions are plotted and given in Fig. \ref{fig1}.
After the transformation of the values to the interval [0, 1], three
methods are used to convert the values to a binary number. These
methods are $x_i(k)=\{\begin{array}{l}
          1, rand < T(x_i(k)) \\
          0, otherwise
          \end{array}$, $x_i(k)=\{\begin{array}{l}
    x^*(k), rand < T(x_i(k)) \\
    0, otherwise
    \end{array}$ and $x_i(k)=\{\begin{array}{l}
                           (x_i(k))^{-1}, rand < T(x_i(k)) \\
                           x_i(k), otherwise
                           \end{array}$
where $x_i(k)$ is the $k^{th}$ component of $x_i$ from the previous
iteration, $x^*$ the best solution so far from the memory and
$(x_i(k))^{-1}$ gives the complement of $x_i(k)$, that is if
$x_i(k)=1$ its inverse will be 0 and if it is 0 its inverse will be
1.

In addition to using sigmoid function for the conversion of the
variables to binary numbers, the updating formula was modified in
\cite{Farhoodnea101, Farhoodnea102}. The distance is modified using
sigmoid function using $S(r)=|tanh(\lambda r)|$ for a parameter
$\lambda$ with value near 1 and based on this the updating formula
is done using $x_i=\{\begin{array}{l}
x_i+\beta(x_j-x_i)+\alpha (rand() - 0.5), rand < S(r) \\
x_i, otherwise
\end{array}$. If firefly $x_i$ is closer $x_j$ then it has less probability of moving.
That may affect the quality of the solution as rather than improving
or exploring the solution will stay in its position. In
\cite{Costa112}, another version of S-shaped sigmoid function given
by $S(x_i(k))=0.5(1+erf()x_i(k))$ is proposed.

In addition to modification for binary problems, firefly algorithm
has also been used for optimization problems with integer valued
variables. In \cite{Bacanin123, Baghlani126}, the same updating
formulas as the standard firefly algorithm used and the result is
rounded to the nearest integer value. A similar approach is used for
the mixed integer problem in \cite{Gandomi125}. Another modification
for binary mixed problem is presented in \cite{Bacanin127}. The
modification is done using a new updating formula given by
$x_i(k):=round(\frac{1}{1+e^{-x_i(k)+rand(x_i-x_j)}}-0.06)$.

Another modification in this category is done in \cite{Oliveira114,
Marichelvam116}. The modification is based on a concept of random
key, which is proposed in \citep{Bean94}. The method uses a mapping
of a random number space, [0, 1], to the problem space. In other
words, it encodes and decodes a solution with real numbers and these
numbers, obtained randomly in a (0,1) uniform probabilistic
distribution, are keys for sorting other numbers in order to form
feasible solutions in an optimization problem and also in the
updating process of firefly algorithm. Hence, after updating a
solution using the standard firefly algorithm updating mechanism
then it will be converted to integers using randem key approach.

\begin{figure}[H]
\centering
\includegraphics[width=12cm, height=7cm]{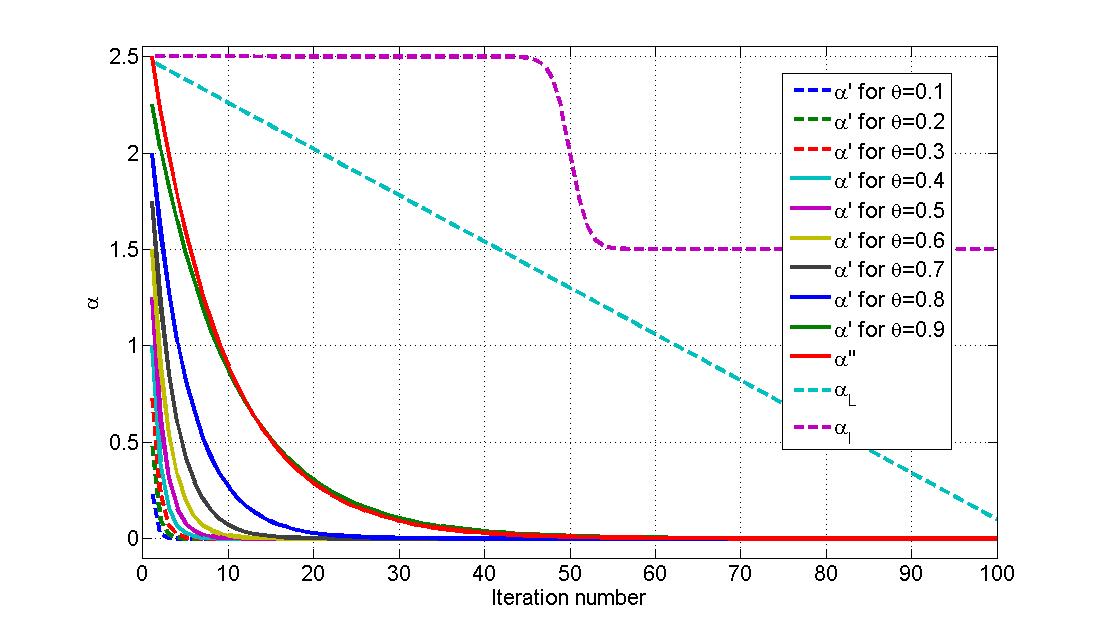}
\caption{Different modifications proposed for $\alpha$, $\alpha''$
is the modification proposed in [127], $\alpha_L$ from [112],
$\alpha_l$ from [108] and $\alpha'$ is from [125, 123, 126] with
different values of $\theta$, for all the cases
$\alpha_0=\alpha_{max}=2.5$ and for the $\alpha_L$
$\alpha_{min}=0.1$.}\label{fig2}
\end{figure}

In addition to modifying the algorithm to suit the problem,
different additional modifications are proposed to increase the
effectiveness of the algorithm. In \cite{Bacanin123, Baghlani126,
Gandomi125}, the random movement step length was made adaptive as a
function of the iteration number, $Itr$, using
$\alpha=\alpha_0\theta ^{Itr}$, for a new algorithm parameter
$\theta$, where $0<\theta<1$. Fig. \ref{fig2} shows the adaptive
step length proposed for different values of $\theta$. Another
modification on the random movement step length is given in
\cite{Bacanin127}, by $\alpha:=\alpha
(\frac{10^{-4}}{9})^{\frac{1}{MaxItr}}$. Additionally, a scaling
parameter, $x_{max}-x_{min}$, multiplies the random movement
\cite{Bacanin123, Baghlani126}. The randomness step length $\alpha$
is modified using
$\alpha=\alpha_0-\frac{1}{1+e^{-(Itr-\frac{MaxItr}{2})}}$, in
\cite{Liu108}. Furthermore, in \cite{Chhikara99}, $\alpha$ and
$\gamma$ is modified based on the problem property.

In \cite{Costa112}, both $\alpha$ and $\gamma$ are made to change
with iteration using
$\alpha=\alpha_{max}-\frac{Itr}{MaxItr}(\alpha_{max}-\alpha_{min})$
and
$\gamma=\gamma_{max}e^{\frac{Itr}{MaxItr}ln(\frac{\gamma_{min}}{\gamma_{max}})}$.
In this modification $\alpha$ decreases linearly and $\gamma$
increases quicker than a linear function. This implies that as the
iteration increases the random movement decreases so does the
attraction step length. In addition Levy distribution is used to
generate a random direction with a scaling parameter which is the
difference between the maximum and minimum values of the feasible
region. However, it should be noted that Levy distribution will
generate a random direction with a step length and also $\alpha$ is
used to control the step length hence adding additional scaling
parameter may not be effective. That is because it is possible to
control the step length based on $\alpha$ and the random vector
generated by Levy distribution. Additionally, since $\alpha$ is made
to decrease through iteration, its effect may not be seen due to
this scaling parameter.

\subsection{Updating in the discrete space}

\begin{figure}[H]
\centering
\includegraphics[width=12cm, height=7cm]{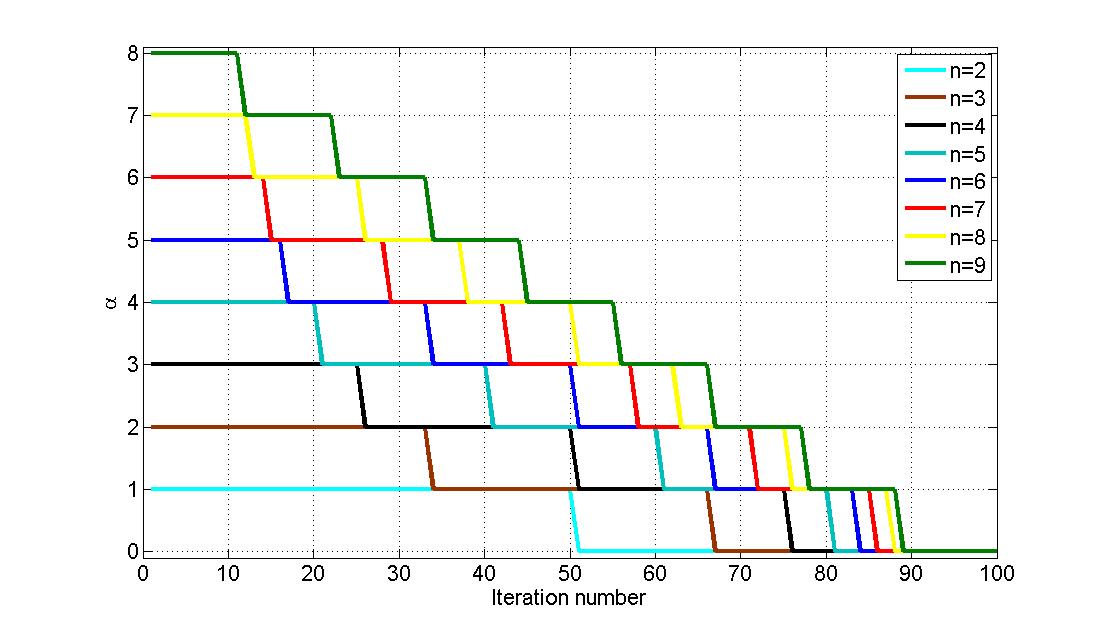}
\caption{The step length $\alpha$ for dimensions of
$n=2,3,4,5,6,7,8,9$ and for 100 iterations}\label{fig3half}
\end{figure}

In \cite{Poursalehi117, Durkota118}, the standard firefly algorithm
is modified for loading pattern enhancement. The generation of
random solutions are using random permutation and the distance
between fireflies are measured using hamming distance. Hamming
distance of two vectors $x_i$ and $x_j$ is given by $d=|H|$ where
$H$ is number of entries, $k$, for which $x_i(k) \neq x_j(k)$. The
updating process is separated and made sequentially;first the
$\beta$-step, a move due to the attraction and $\alpha$-step, a move
due to the random movement. In the $\beta$-step, first same entries
with same index for both fireflies, $x_i$ and $x_j$ will be
preserved and for the other components an entry from $x_j$ will be
copied if $rand<\beta$, where $\beta=\frac{1}{1+\gamma d^2}$. If
$rand\geq \beta$, then the entry from $x_i$ will be used. After
moving $x_i$ using the $\beta$-step the random movement or the
$\alpha$-step will be used to update $x_i$ using
$x_i:=round(x_i+\alpha (rand()-0.5))$ with a swapping mechanism to
preserve feasibility. A similar approach, but different way of
computing $\beta$ is proposed in \cite{Ishikawa119}. It is computed
based on the familiarity degree $P$, which is a random $N$ by $N$
vector initially and updates by
$P_{ij}=P_{ij}+\frac{1}{|rank_i-rank_j|}$ and the
$\beta=e^{-\frac{(\max\limits_k\{P_{ik}\}-P_{ij})^2}{\max\limits_k\{P_{ik}\}}}$.
In addition to making the algorithm suitable for non-continuous
variables, in \cite{Ishikawa119}, the randomness parameter $\alpha$
is made adaptive using $\alpha=\lfloor
n-\frac{Itr}{MaxItr}n\rfloor$, which is a function of the iteration
number and the dimension of the problem. As can be seen from
Fig.\ref{fig3half}, it decreases with iteration.

The modification proposed in \cite{Poursalehi120} is similar with
the modification proposed in \cite{Poursalehi117, Durkota118}. The
only difference is in the updating mechanism, which is done based on
a given probability parameter $\rho$,
$\rho=0.5+\frac{0.5Itr}{MaxItr}$. It is the probability for a
firefly $x_i$ to follow another brighter firefly $x_j$. However, if
$rand > \rho$ firefly $x_i$ will move towards the brightest firefly
of the fireflies.

In \cite{Mamaghani113}, after a randomly integer coded initial
solutions are generated, hamming distance is used to compute the
distance, $r$, between two solutions. Then a random number $R$ will
be generated between 1 and $r$, and $R$ swaps will be done from the
brighter firefly. A similar modification is proposed in
\cite{Osaba200}. In \cite{Osaba200}, after computing the humming
distance, $r$, of two fireflies a random number $R$ between 2 and
$r\gamma^{Itr}$ will be used to in stead of $r$ in
\cite{Mamaghani113}. Only improving solutions will be accepted.

For travel salesman problem, firefly algorithm has been modified in
\cite{Jati121}. The distance between two solution is computed using
$r=10\frac{A}{n}$, where $A$ is number of different arcs and $n$ is
the number of cities. A firefly $i$ moves towards brighter firefly
$j$, $m$ times where $m$ is a new parameter to determine the number
of moves, using inversion mutation to preserve feasibility. That is
an initial chromosome is selected randomly and other entries will be
filled using inversion mutation $m$ times to generate $m$ new
solutions. Once all the $N$ solutions are moved $m$ times then the
best $N$ solutions will be selected to pass to the next iteration.

\begin{figure}[H]
\centering
\includegraphics[width=12cm, height=7cm]{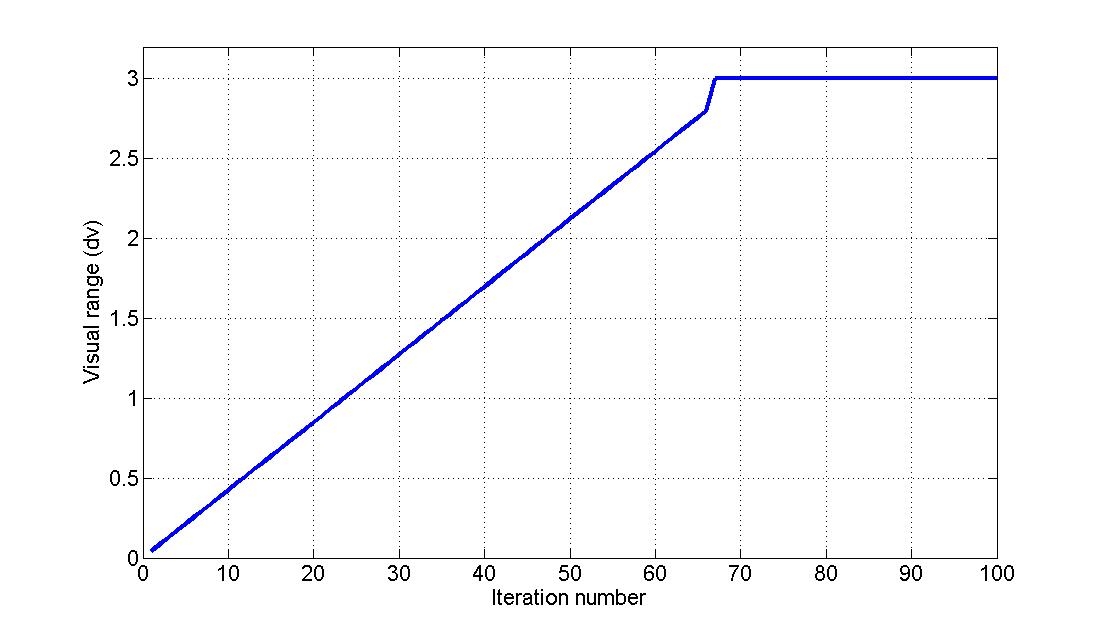}
\caption{The visual range of a firefly with $dv_{max}=3$ and
$dv_{min}=0.2$ for 100 iterations}\label{fig3}
\end{figure}

Another modification is proposed in \cite{Li122}. The updating for a
solution $x_i$ towards a brighter firefly $x_j$ is given by
$x_i(k):=\{\begin{array}{l}
S_i(k), \alpha |rand-0.5| < \beta_0e^{-\gamma r^2} \\
x_i(k), otherwise
\end{array}$, where
$S_i(k)=\{\begin{array}{l}
x_j(k), x_j(k)\neq x_i(k) \\
0, otherwise
\end{array}$, for each dimension $k$. In addition to this,
a firefly $x_i$ will be affected by fireflies in its visual range.
That means for a firefly $x_i$ to move towards another firefly
$x_j$, firefly $x_j$ should be brighter and also should be in
$x_i$'s visual range. The visual range, $dv$, is computed using
$dv=\{\begin{array}{l}
\frac{3(dv_{max}-dv{min})Itr)}{2(MaxItr-1)}, Itr<\frac{2MaxItr}{3} \\
dv_{max}, otherwise
\end{array}$. As can be seen from Fig. \ref{fig3}, the visual range
increases until it reaches to the maximum value. This means that
initially a brighter firefly affects solution near its location
where in latter stages it can affect any firefly in the maximum
allowed visual range.

A  modification based on the problem property for supplier selection
problem is also proposed in \cite{Kota124}. In addition in
\cite{Baykasoglu110}, firefly algorithm algorithm has been modified
and used for knapsack problems. The discretization is done based on
the problem property. In addition, a firefly $i$ will move towards a
brighter firefly $j$ if $rand <
rank_i^{-\frac{mode(Itr-1,MaxItr)}{MaxItr}}$. The additional
condition slowly vanishes as the iteration increases. Furthermore,
$\beta$ is modified for computational reason using
$\beta=\frac{\beta_0}{\omega+r}$, for a very small number $\omega$
to omit the singularity case. A similar modification is used in
\cite{Baykasoglu111}. In addition to the discretization done in
\cite{Baykasoglu110}, the authors in \cite{Baykasoglu111} proposed
two additional moves after the updates. The first one is a random
flight by 10\% of top fireflies with 0.45 probability. The move will
be accepted only if it is improving. The second is a local search of
the brighter firefly. After 10\% of the iterations it will do a
local search and the update will be accepted if it is improving.

\section{Discussion}

\begin{figure}[H]
\centering
\includegraphics[width=10cm, height=6cm]{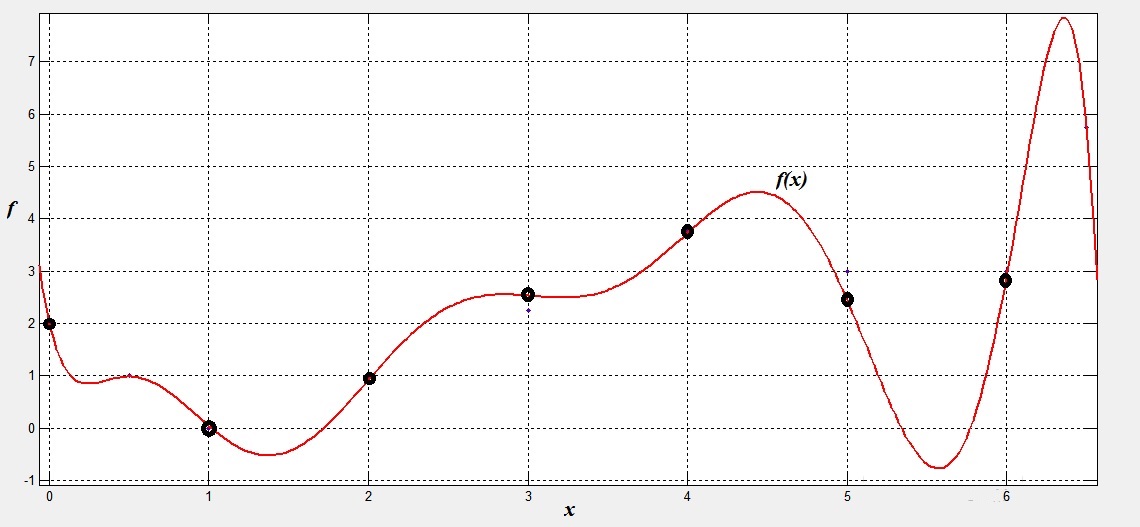}
\caption{Discrete versus continuous space search where the black
circles representing the ineteger functional value and the red graph
is the corresponding continuous function}\label{Fig4}
\end{figure}

Binary problems are one of the main class of problems with
non-continuous variables for which firefly algorithm has been
modified for. Most of these modifications deals with updating the
continuous space using the updating formula of the standard firefly
algorithm and changing the resulting solution to a binary number for
each dimension. The conversion usually uses sigmoid functions which
will convert the result to be in between zero and one. The decimal
number then converted to a binary number. Updating the solutions
using the updating formula of the standard firefly algorithm and
converting the result is done not only for binary problems but also
for other problems with non-continuous variables. However, it should
be noted that updating on the continuous space push the solutions to
the continuous solution which can be far away from the solution for
the discrete variables. For example if we consider the function
given in Fig. \ref{Fig4}, the solution for the continuous case is
around $x=5.5$ and the nearest integer is $x=5$ and $x=6$. However,
for the integer valued problem the solution is at $x=1$. Hence,
updating on the discrete space has an advantage.

In updating in the discrete space usually done based on the coding
used which is based on the properties of the problems. The distance
between two fireflies are measured based on the difference in the
entries or based on the difference sequence of the entries. Swapping
using different approaches are used. A research on which approach is
better has not be studied and can be studied in the future.
Modifying an algorithm to suit a given problem has also been done
and it is often effective. Another possible future work is
generalizing the approach so that a given modification in one
problem domain can be used in another.

Apart from making the algorithm suitable for problems with
non-continuous variables the algorithm parameters are modified.
Making the randomness step length $\alpha$ decreasing with iteration
is a good idea in order to archive quality solutions. In addition it
has also been modified based on the problem dimension however
appropriate scaling parameter needs to accompany the modification.
Perhaps it is also an ideal  issue to explore the adaptive step
length which varies based on the performance and also which can
possibly be increasing and also decreasing based on its current
status.

Another possible future work is to merge the search mechanism both
on the discrete as well as the continuous space. Perhaps the
advantage of the search on continuous space and also discrete space
can be combined to give a superior performance.

\section{Conclusion}

Firefly algorithm has been proposed for continuous problems.
However, due to the application of optimization problems with
non-continuous problems, it has been modified and used in different
studies. The modification basically can be categorized into two. The
first category is where the updating mechanism is done on the
continuous space and the result is converted to the discrete values.
For this purpose different sigmoid and tan hyperbolic functions are
used most of the time. The second category is when the update is
performed on the discrete space. This approach has an advantage over
the first as it will may lead the solutions in a local minimum or
wrong direction. In this paper, a review on these modifications
along with possible future works is discussed.

\bibliography{mybibfile}

\end{document}